%% file: main.tex
\documentclass{article}
\usepackage{spconf,amsmath,graphicx}
\usepackage{color}
\usepackage{amssymb}
\usepackage{multicol,multirow}
\usepackage[compatibility=false, font=footnotesize]{caption}
\usepackage{xspace}
\usepackage[numbers,sort&compress]{natbib}
\usepackage{setspace}

\title{CM$\boldsymbol{^2}$-Net: Continual Cross-Modal Mapping Network for Driver Action Recognition}
\name{Ruoyu Wang$^{\dagger}$, Chen Cai$^{\dagger}$, Wenqian Wang$^{\star}$, Jianjun Gao$^{\dagger}$, Dan Lin$^{\star,\ddagger}$, Wenyang Liu$^{\dagger}$, Kim-Hui Yap$^{\dagger}$}
\address{
$^{\dagger}$School of Electrical and Electronic Engineering, Nanyang Technological University, Singapore\\
$^{\star}$ Continental-NTU Corporate Lab, Nanyang Technological University, Singapore\\
$^{\ddagger}$ College of Computer Science and Technology, Harbin Engineering University, China
}
\begin{document}
%
\maketitle

\input{Sec/0_abstract_3}

\input{Sec/1_intro}

\input{Sec/2_related}
\input{Sec/3_methodology}
\input{Sec/4_experiments}
\input{Sec/5_conclusion}


\begin{spacing}{0.9}
\setlength{\bibsep}{0.3em}
\bibliographystyle{unsrt}
\bibliography{ref}
\end{spacing}

\end{document}

%% file: Sec/0_abstract_3.tex
\begin{abstract}
Driver action recognition has significantly advanced in enhancing driver-vehicle interactions and ensuring driving safety by integrating multiple modalities, such as infrared and depth.
Nevertheless, compared to RGB modality only, it is always laborious and costly to collect extensive data for \textbf{\textit{all}} types of non-RGB modalities in car cabin environments. 
Therefore, previous works have suggested \textbf{\textit{independently}} learning each non-RGB modality by fine-tuning a model pre-trained on RGB videos, but these methods are less effective in extracting informative features when faced with newly-incoming modalities due to large domain gaps.
In contrast, we propose a Continual Cross-Modal Mapping Network (CM$^2$-Net) to \textbf{\textit{continually}} learn each newly-incoming modality with instructive prompts from the previously-learned modalities. 
Specifically, we have developed Accumulative Cross-modal Mapping Prompting (ACMP), to map the discriminative and informative features learned from previous modalities into the feature space of newly-incoming modalities.
Then, when faced with newly-incoming modalities, these mapped features are able to provide effective prompts for which features should be extracted and prioritized. 
These prompts are accumulating throughout the continual learning process, thereby boosting further recognition performances.
Extensive experiments conducted on the Drive\&Act dataset demonstrate the performance superiority of CM$^2$-Net on both uni- and multi-modal driver action recognition.

\end{abstract}
\begin{keywords}
Driver action recognition, cross-modal learning, continual learning
\end{keywords}

%% file: Sec/1_intro.tex
\vspace{-10pt}

\section{Introduction}

\vspace{-5pt}

\begin{figure}[t]
\centering
\includegraphics[width=0.48\textwidth]{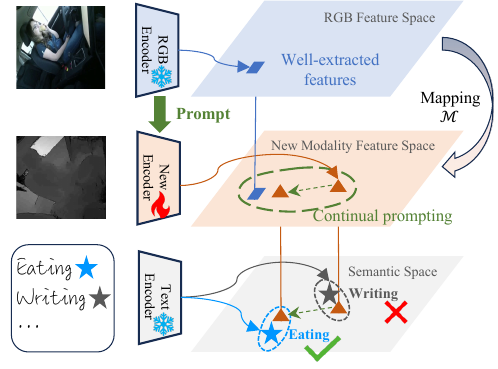}
\caption{
The features extracted by pre-trained encoders (such as RGB) are discriminative and informative~\cite{girdhar2023imagebind}.
Instead of training a new encoder from scratch without any prior knowledge, we propose to map the well-extracted features into the new modality feature space to prompt the training.
The prompt assists in aligning the features extracted by the new encoder with textual embeddings representing driver actions in the semantic space, thereby enhancing the robustness and accuracy of driver action recognition.
}
\label{motivation}
\vspace{-12pt}
\end{figure}

Recent technological breakthroughs in autonomous driving have significantly increased flexibility for drivers within car cabins.
Therefore, employing automatic driver action recognition through various in-car monitoring sensors is crucial for enhancing driver-vehicle interaction and ensuring driving safety.
Current approaches~\cite{dar, mod4, khan2022supervised} integrate various vision modalities, such as RGB, InfraRed (IR), and Depth videos, captured through complementary sensors, to accurately recognize diverse in-car human behaviors.

These methods~\cite{mod4, khan2022supervised} employ extensive RGB data pre-trained backbone networks~\cite{clipvit, swin, uniformerv2} to extract features from various modalities and fuse them to perform multi-modal driver action recognition.
But it presents a challenge in extracting informative features from IR or Depth videos using such RGB-based pre-trained backbones~\cite{dar}, given the existence of inherent gaps between RGB and other modalities. 
Extracting accurate multi-modal visual features using the respective well-pre-trained backbone networks may bypass this problem. 
However, to the best of our knowledge, there is a lack of large-scale IR or Depth-specific pre-trained models capable of accurately extracting features for these modalities.
This limitation potentially compromises driver action recognition accuracy, especially in scenarios where RGB is unreliable, such as at midnight or inside tunnels. 
In such situations, it is essential to collect informative visual features from other modalities to assist the model in accurate recognition.
In this study, we aim to transfer the knowledge from well-pre-trained encoders to other novel or less-explored modalities, thereby achieving continual learning of new modalities and enhancing their feature extraction capabilities for better driver action recognition.
In the process of feature extraction, which can be characterized as a form of lossy mapping, the issue with inefficient encoders lies in their potential failure to retain essential information.
This loss of essential information significantly impacts the subsequent analytical and predictive tasks for which the features are used.
Inspired by ImageBind~\cite{girdhar2023imagebind}, which investigated the use of RGB data to align different modalities into a unified feature space, we find that RGB-derived features are not only informative and comprehensive but also capable of navigating the learning process of other modalities.
Therefore, we can boost the encoding ability by offering targeted prompts such as RGB-derived features to guide these modalities' specified encoders on essential information to retain.
As depicted in Fig.~\ref{motivation}, we employ well-trained RGB-based encoders for prompting during the training of the new modality encoder. 
Upon completing each training phase, the newly-trained modality encoder is able to continually facilitate the training of subsequent encoders and bridge the domain gap of the modality features for better driver action recognition.

To solve the aforementioned challenges, we present a novel Continual Cross-Modal Mapping Network (CM$^2$-Net), which leverages the knowledge from existing well-pre-trained models to learn each new modality, enabling the extraction of reliable multi-modal features for accurate driver action recognition.
We initially fine-tune a pre-trained RGB-based encoder (e.g., UniFormerV2~\cite{uniformerv2}), leveraging its strong generalization and representation learning abilities to encode RGB features.
Subsequently, we explore an Accumulative Cross-modal Mapping Prompting (ACMP) strategy to prompt the training of other modalities encoders.
Specifically, we utilize our newly-tuned RGB encoder to extract discriminative RGB features and map them to continually prompt the feature-learning process of other modality-specific encoders.
In addition, each modality learned with ACMP also contributes to prompt the training of subsequent modality-specific encoders.
Through this learning process, the newly-trained modality-specific encoder can leverage the expertise of previously well-trained encoder to effectively encode modality-specific features, resulting in improved feature representation learning.
Furthermore, we employ a frozen language encoder (e.g., CLIP~\cite{clipvit}) to encode label textual knowledge into embeddings for stable and consistent supervisory signals.
The encoded textual features can function as pivots, guiding the learning of multimodal features within a unified feature space during continual training of different modality-specific encoders. 
Additionally, the semantic knowledge is injected into multi-modal features, thereby enhancing action recognition.

Our contributions are summarized below. 
(1) We introduce the CM$^2$-Net, a cross-modal driver action recognition model, specifically presented to adaptively and continually learn from new modalities.
(2) We have devised the Accumulative Cross-modal Mapping Prompting (ACMP) method, which utilizes the knowledge and information of our previously-learned modalities to facilitate the feature learning of new modalities.
(3) Our proposed method demonstrates state-of-the-art performance in both uni- and multi-modal driver action recognition tasks on the challenging Drive\&Act dataset.

%% file: Sec/2_related.tex
\vspace{-10pt}

\section{Related Work}

\vspace{-5pt}

\subsection{Driver Action Recognition}

\vspace{-5pt}

Driver action recognition, which aims to identify behaviors of a driver in the car cabin, is different from generalized action recognition task~\cite{i3d, dynamicAR, tsm, maginet, c3d,wang2022ae}.
This task is challenging due to the unstable lighting conditions in the car cabin.
Lin et al.~\cite{dar} proposed an efficient multi-modal model DFS to address this limitation of unimodal methods.
Rotiberg et al.~\cite{mod4} conducted experiments to comparatively evaluate the performance of different decision-level fusion methods across multi-modality inputs.
Although these approaches employed various encoders to extract multi-modal features respectively for the final prediction, they did not address the limitations of their performance in comparison to RGB encoders.
In contrast, our CM$^2$-Net aims to transfer the robust feature extraction capabilities and knowledge of well-pre-trained encoders to the new modal-specific encoders for more accurate recognition of driver behaviors.


\begin{figure*}[ht]
\centering \vspace{-2pt}
\includegraphics[width=1.0\textwidth]{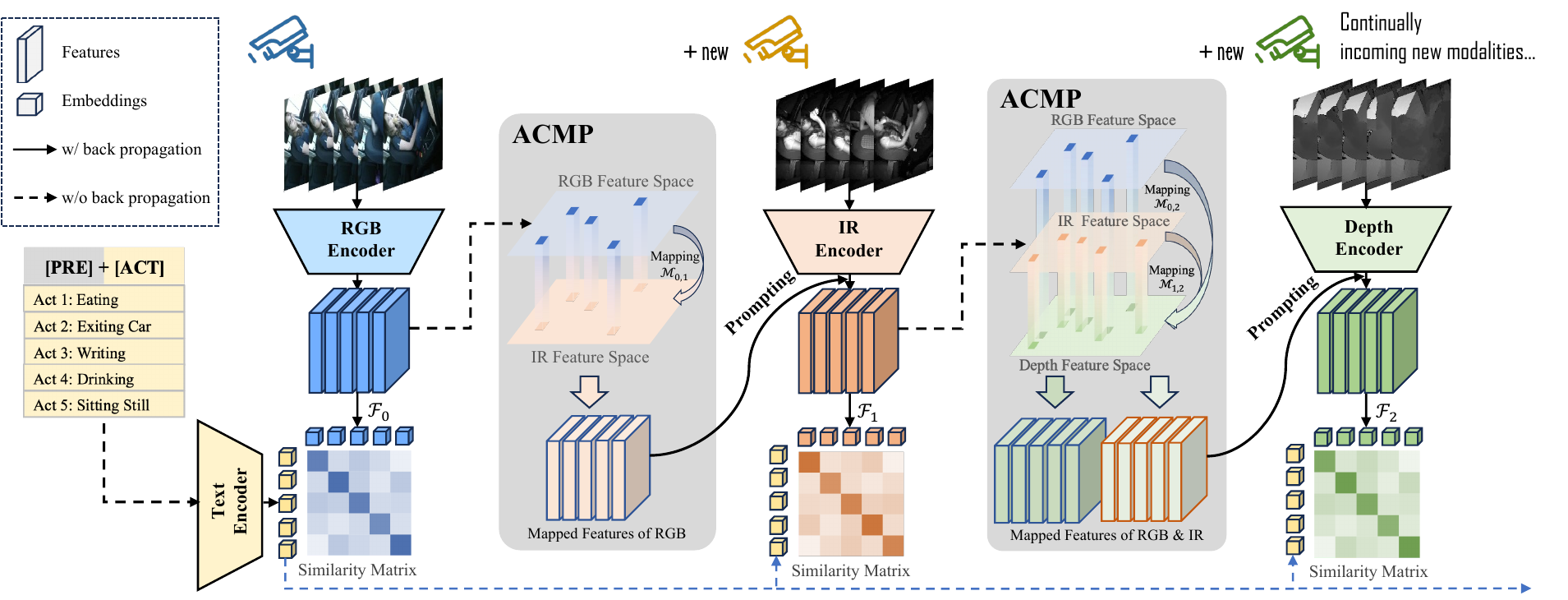}
\caption{
\textbf{Overview} of CM$^2$-Net, a network designed for continual learning across different modalities in driver action recognition.
Initially, CM$^2$-Net begins by fine-tuning an RGB encoder to learn discriminative RGB features and classifies the driver actions based on the similarity scores with label textual embeddings.
Then, for a new modality (such as Depth), CM$^2$-Net employs Accumulative Cross-modal Mapping Prompting (\textbf{ACMP}) to train a modality-specific encoder (such as Depth Encoder) with the prompting from previously-learned modalities.
ACMP can effectively map the accumulating discriminative features from multiple established modalities (such as RGB and IR) into the new modality feature space, prompting which crucial features should be extracted during encoder training.
In this way, the prompts can improve the alignment between the new modality embeddings and textual embeddings for accurate classification.}
\label{overview} \vspace{-10pt}
\end{figure*}

\vspace{-8pt}

\subsection{Cross-modal Learning}

\vspace{-5pt}

Recently, with the rapid advancement of large language models, cross-modal learning has garnered significant attention in the field of computer vision~\cite{TDcap, girdhar2023imagebind,cai2022attribute, wang2022multi}.
Radford et al.~\cite{clipvit} proposed CLIP to align image and text inputs within a shared semantic space, enabling effective cross-modal understanding and zero-shot learning capabilities.
In ImageBind, Girdhar et al.~\cite{girdhar2023imagebind} employed contrastive learning to pair the image and other modalities in a common space and achieved great success in both retrieval and generation tasks.
Different from these approaches, CM$^2$-Net focuses on employing cross-modal learning to enhance the model representation capability.

\vspace{-8pt}

\subsection{Continual Learning}

\vspace{-5pt}

Continual learning aims to mimic human intelligence by progressively acquiring new skills in real-world scenarios~\cite{continualsurvey,else-net}. 
The typical one focuses on the class dimension, learning from evolving data distributions while retaining knowledge of previously encountered data~\cite{else-net, pivot}.
Inspired by the learning paradigm of these works, we put forward to our modal continual learning, which is to continually learn new modalities.
CM$^2$-Net is able to accumulate the prompts during learning each new modality, thereby enhancing generality in multi-modal scenarios.

%% file: Sec/3_methodology.tex
\vspace{-5pt}

\section{Methodology}

\vspace{-5pt}

In this section, we present \textbf{CM$^2$-Net} - a \textbf{C}ontinual \textbf{C}ross-\textbf{M}odal \textbf{M}apping \textbf{Net}work for driver action recognition.
As shown in Fig.~2, CM$^2$-Net is able to continually learn new modalities with Accumulative Cross-modal Mapping Prompting (ACMP).
Given that features extracted by pre-trained encoders are comprehensive and informative, ACMP can map them into the new modality feature space to prompt the new encoder for effective training.
In initialization, we first fine-tune a well-pre-trained RGB backbone under supervision from label text embeddings.
Subsequently, when faced with a newly-incoming modality, CM$^2$-Net leverages previously-extracted salient features to prompt the training of the new encoder.
Thus, the prompts is accumulating during the continual learning of new modalities and can be utilized to prompt the training of next new modality.
CM$^2$-Net achieves self-evolution in multi-modal scenarios and improves the accuracy in both uni- and multi-modal recognition.

\vspace{-8pt}

\subsection{Text to RGB as Initialization}

\vspace{-5pt}

In initialization of CM$^2$-Net, we select RGB as the first modality to be learned for the strong generalization capability of the pre-trained RGB backbone.
CM$^2$-Net loads a pre-trained model to initialize RGB encoder and then employs a frozen text encoder to provide target embeddings for fine-tuning in semantic space.
Unlike previous action recognition methods, which map the visual input into discrete numerical sets, CM$^2$-Net aims to embed the visual into a semantic latent space, classifying them based on similarity scores with label text embeddings.
In such a way, it not only learns the semantic information but also identifies a robust classification space in the training of other modalities.

For the RGB video clip input $\boldsymbol{x}_0$, we utilize the corresponding video encoder $\mathcal{V}_0$ to extract its spatiotemporal features.
Accordingly, the text encoder $\boldsymbol{T}$ is utilized to represent label text $\boldsymbol{y}$ into textual features with the prefix prompt \texttt{[PRE]}.
Here, we set \texttt{[PRE]} to "\textit{A video of a driver }" to contextualize the text.
The text encoder $\mathcal{T}$ can be adopted from any well-designed language model.
Projection heads $\mathcal{F}_0$ and $\mathcal{F}_t$, consisting of multilayer perceptions, are then utilized to embed these features into the semantic latent space.
\begin{equation}\label{embedding}
    \mathbf{v}_0=\mathcal{F}_0\mathcal{V}_0(\boldsymbol{x}_0), \quad \mathbf{w}=\mathcal{F}_t\mathcal{T}(\texttt{[PRE]} \oplus\boldsymbol{y}).
\end{equation}
where $\oplus$ denotes concatenation operation.
By representing both video and label text as embeddings within the unified semantic space, the encoder is trained by contrastive learning, bringing embeddings from the same class closer, while distancing those from different classes. 
We calculate the similarity score between video and label text using the cosine similarity between the embeddings:
\begin{equation}\label{similarity}
    \boldsymbol{S}(\boldsymbol{x}_0, \boldsymbol{y}) = \mathbf{cos}(\mathbf{v}_{\boldsymbol{0}}, \mathbf{w})
\end{equation}
where $\boldsymbol{S}$ is used as a classification score in the further inference and classification loss $\mathcal{L}_{cls}$ calculation.
After training, this RGB encoder is frozen and employed to prompt the training of following modalities.

\vspace{-8pt}

\subsection{Accumulative Cross-modal Mapping Prompting}

\vspace{-5pt}

In this subsection, we first demonstrate that well-extracted features contain information that is shared across modalities for the same action. 
It can be transferred into different modalities with the mapping function $\mathcal{M}$.
Building on this concept, we introduce our ACMP, which is able to map the shared information conducted from previously-learned modalities into the new modality feature space to prompt the training of a new encoder.
In this way, we can leverage the model's cross-modal knowledge into the new encoder and enhance its feature extraction and representation capability.

Considering that data from different modalities capture the same action, we find that there is underlying information shared across these modalities.
For easy demonstration, we assign an index value of $0$ to the RGB modality within these mappings in the following.
For an action $\boldsymbol{X}$ in the real scenario, the different sensors (such as RGB camera and IR camera) utilize different lossy mappings $\phi_i(\boldsymbol{\cdot})$ to capture it into different modality data $\boldsymbol{x}_i$:
\begin{equation}
    \boldsymbol{x}_i = \phi_i(\boldsymbol{X})
\end{equation}
where $i$ denotes the $i$-th modality.
Meanwhile, feature extraction in the encoder is another lossy mapping process $\mathcal{V}_i$ of the modality data $\boldsymbol{x}_i$.
\begin{equation}
    \boldsymbol{f}_i = \mathcal{V}_i(x_i)
\end{equation}
A primary objective of an effective encoder is to minimize the loss of discriminative information during feature extraction. 
To achieve this, the training of the encoder is focused on preserving this crucial information. 
Previous studies, such as ImageBind~\cite{girdhar2023imagebind}, have demonstrated significant success in pairing RGB input with other modality inputs in latent space. 
Building upon this foundation, there exists a series of projection functions $\mathcal{F}_{0, i}$ able to map the different modality features into the same latent space and preliminary results suggest that the information retained in this latent space is sufficiently discriminative to enable effective classification.
So with powerful enough encoders $\mathcal{V}_i$, we can have
\begin{equation}
    \mathcal{F}_0\mathcal{V}_0\phi_0(\boldsymbol{\cdot}) = \mathcal{F}_{0,i}\mathcal{V}_i\phi_i(\boldsymbol{\cdot}), \forall i = 1,2,\boldsymbol{\cdots}
\end{equation}
This equation signifies that discriminative information, shared across modalities, is retained through multiple lossy mappings, and we denote it as $\mathrm{H}_0(\boldsymbol{\cdot})$.
\begin{equation}
    \mathrm{H}_0(\boldsymbol{\cdot}) = \mathcal{F}_0\mathcal{V}_0\phi_0(\boldsymbol{\cdot})
\end{equation}

As critical information $\mathrm{H}_0$ shared across modalities, the objective of ACMP is to prompt each modality encoder to extract and preserve the discriminative information.
Thus, it is essential to map it into each new modality feature space.
For a new modality, if and only if the projection head is bijective, then the inverse function of $\mathcal{F}_{0, i}$ exists and we denote it as $\mathcal{F}_{0,i}^{-1}$.
The inverse function $\mathcal{F}_{0, i}^{-1}$ can map all of the information into the new modality feature space.
\begin{equation}     
\mathcal{F}_{0,i}^{-1}\mathrm{H}_0(\boldsymbol{\cdot})=
\mathcal{F}_{0,i}^{-1} \mathcal{F}_0\mathcal{V}_0\phi_0(\boldsymbol{\cdot}) = \mathcal{F}_{0,i}^{-1} \mathcal{F}_{0,i}\mathcal{V}_i\phi_i(\boldsymbol{\cdot}) = \mathcal{V}_i \phi(\boldsymbol{\cdot}) 
\end{equation}
In cases where the projection head is not bijective, we can construct a mapping function $\Theta_{0,i}(\boldsymbol{\cdot})$ to translate part of the information represented by $\mathrm{H}_0(\boldsymbol{\cdot})$ back into the feature space.
\begin{equation}
    \Theta_{0,i}\mathrm{H}_0(\boldsymbol{\cdot}) = \Theta_{0,i}\mathcal{F}_0\mathcal{V}_0\phi_0(\boldsymbol{\cdot}) =
    \Theta_{0,i}\mathcal{F}_{0,i}\mathcal{V}_i\phi_i(\boldsymbol{\cdot}) \subseteq \mathcal{V}_i\phi_i(\boldsymbol{\cdot})
\end{equation}
To note, our discriminative RGB features are represented by $\mathcal{V}_0\phi_0(\boldsymbol{\cdot})$.
Thus, in the training of the new encoder $\mathcal{V}_i$, the objective mapping function to map RGB features is $\Theta_{0,i}\mathcal{F}_0(\boldsymbol{\cdot})$ and we denote is as $\mathcal{M}_{0,i}(\boldsymbol{\cdot}) = \Theta_{0,i}\mathcal{F}_0(\boldsymbol{\cdot})$.
In CM$^2$-Net, we train a decoder consisting of projection heads as $\mathcal{M}_{0,i}(\boldsymbol{\cdot})$.
During training, we utilize contrastive learning to compare the mapped information $\mathcal{M}_{0,i}\mathcal{V}_0\phi_0(\boldsymbol{\cdot})$ with the extracted features so that this part of knowledge in RGB encoder can be leveraged into the new encoder.

The information in prompts is accumulating with the continual learning of modalities.
After training, the new encoder $\mathcal{V}_i$'s capability to extract features of the $i$-th modality is significantly enhanced.
ACMP also learns a mapping function $\mathcal{M}_{i,j}$ to map the shared information from the $i$-th modality into the following modality $\boldsymbol{x}_j$.
Thus, ACMP is able to make use of the accumulating information from all previously extracted features for prompting.
The total information that can be delivered to the new encoder $\mathcal{V}_{j}$ is:
\begin{equation}
    \mathbf{H}_{j}=\bigcup_{i=0}^{j-1}\Theta_{i,j}^{-1} \mathrm{H}_i=\bigcup_{i=0}^{j-1} \Theta_{i,j}^{-1} \mathcal{F}_i \mathcal{V}_i \phi _i = \bigcup_{i=0}^{j-1} \mathcal{M}_{i,j}\mathcal{V}_i \phi_i
\end{equation}
It is important to emphasize that the information mapping in ACMP exclusively occurs during the training of the new encoder and is intended to prompt the training.

\vspace{-8pt}

\subsection{Training}

\vspace{-5pt}

In this subsection, we present our training paradigm and loss function.
When training the $m$-th visual encoder $\mathcal{V}_m (m > 0)$, we keep the text encoder $\mathcal{V}_t$ and other visual encoders$\{\mathcal{V}_i\}_{i=0}^{m-1}$ frozen.
The loss function consists of three parts.
\begin{equation}
    \mathcal{L} = (1-\lambda) \boldsymbol{\cdot} \mathcal{L}_{cls} + \lambda \boldsymbol{\cdot} \mathcal{L}_{text} + \epsilon \boldsymbol{\cdot} \mathcal{L}_{prompt}
\end{equation}
The classification loss $\mathcal{L}_{cls}$ is based on the cross-entropy loss and takes the similarity scores $\boldsymbol{S}$ from Eq.~\ref{similarity} as input.
Apart from $\mathcal{L}_{cls}$, the other two losses are both contrastive losses based on the calculation formula $\boldsymbol{CL}$.
The contrastive loss between label text $\mathcal{L}_{text}$ is calculated as:
\begin{equation}
    \mathcal{L}_{text}=\boldsymbol{CL}(\mathbf{w},\mathbf{v}_m)
\end{equation}
where $\mathbf{w}, \mathbf{v}$ are calculated as Eq.~\ref{embedding}.
For the prompting loss, we contrast features $\boldsymbol{f}_m$ extracted by the new encoder $\mathcal{V}_m$ with the mapping of other modalities features $\mathcal{M}_{i,m}(\boldsymbol{f}_i)$, employing a series of scaling parameters $\omega_i$.
\begin{equation}
    \mathcal{L}_{prompt}={\sum_{i=0}^{m-1}}\omega_i \boldsymbol{\cdot}\boldsymbol{CL}(\mathcal{M}_{i,m}(\boldsymbol{f}_i),\boldsymbol{f}_m)
\end{equation}
Unlike contrastive learning with label text $\mathcal{L}_{text}$ which occurs at the embedding level, the prompt loss $\mathcal{L}_{prompt}$ arises at the feature level.
Here, we take as an example to demonstrate the formula of $\boldsymbol{CL}$ and the prompt loss is calculated similarly.
\begin{equation}
    \mathcal{L}_{text}=\boldsymbol{CL}(\mathbf{w},\mathbf{v}) = \mathcal{L}_{t2v}(\mathbf{w},\mathbf{v}) + \mathcal{L}_{v2t}(\mathbf{w},\mathbf{v})
\end{equation}
\begin{equation}
\begin{split}
    \mathcal{L}_{v2t}(\mathbf{w},\mathbf{v}) = -\sum_{i=1}^{\mathcal{B}} \log\frac{\exp\left(\mathbf{v}_i  \mathbf{w}_{\theta(i)}^T/{\tau}\right)}{\sum_{k=1}^{\mathcal{B}} \exp\left(\mathbf{v}_i t \mathbf{w}_k^T/ {\tau}\right)}\\
    \mathcal{L}_{t2v}(\mathbf{w},\mathbf{v}) = -\sum_{j=1}^{\mathcal{B}} \log\frac{\exp\left(\mathbf{v}_{\varphi(j)} \mathbf{w}_j^T/{\tau}\right)}{\sum_{k=1}^{\mathcal{B}} \exp\left(\mathbf{v}_k \mathbf{w}_j^T/ {\tau}\right)}
\end{split}
\end{equation}
where $\boldsymbol{w}_{\theta(i)},\boldsymbol{w}_{j}$ is the text embedding of the label corresponding to video embedding $\boldsymbol{v}_i, \boldsymbol{v}_{\varphi(j)}$ and $\mathcal{B}$ represents the batch size.
We follow classic method~\cite{clipvit} to compute it as the sum of two parts, the loss of video to text $\mathcal{L}_{v2t}$ and the loss of text to video $\mathcal{L}_{t2v}$.

%% file: Sec/4_experiments.tex
\vspace{-10pt}

\section{Experiments}

\vspace{-5pt}

\subsection{Datasets and Metrics}

\vspace{-5pt}

The Drive\&Act~\cite{driveact} is a comprehensive multi-modal video dataset for driver action recognition. 
This dataset provides five types of multi-modal data: RGB, IR, Depth, Near-InfraRed (NIR), and 3D skeleton data collected from six different views in real driving scenarios.
It categorizes activities into three distinct levels: scenarios, fine-grained activities, and atomic action units.
For the scope of this paper, our experiments are focused on classifying RGB, IR, and Depth videos taken from the right-top view into their respective fine-grained activities, which have 34 classes.
We adhere to the three predefined splits from the dataset to ensure a robust evaluation process and average the outcomes to obtain the final results.

To evaluate our results, two metrics are utilized: Top-1 Accuracy (Top-1) and Mean-1 Accuracy (Mean-1). 
Top-1 assesses the proportion of instances where the top prediction is correct, focusing on the model's highest probability prediction accuracy
Mean-1 averages the accuracies across all classes, ensuring uniform contribution from each and counteracting biases in imbalanced datasets.

\vspace{-8pt}

\subsection{Implementation Details}

\vspace{-5pt}

In the CM$^2$-Net model, we utilize a frozen ViT-B CLIP text model~\cite{clipvit} to encode each label text with prefix into $512$-dimensional features.
For video input, we first sample $8$ frames, resizing each to $224\times224$. 
Then these frames are subdivided into $8\times16\times16$ patches for embedding. 
After that, we employ UniFormerV2-B~\cite{uniformerv2} as video encoders to extract $768$-dimensional features of it. 
Both text and video features are embedded into a $256$-dimensional unified latent space by projection heads for similarity calculation.
We employ contrastive learning with a fixed temperature of $\tau=1.0$ and a loss scaler $\lambda = 0.5$ for the video supervisor contrastive loss. 
During training, we use AdamW~\cite{adamw} as the optimizer.
We set an initial learning rate of $1\times10^{-5}$ and employ a cosine learning rate decay strategy for $100$ epochs.


\vspace{-8pt}

\subsection{Comparison with State-of-the-art}

\vspace{-5pt}

\input{tabs/uni_modal.tex}

We first conduct a comparative analysis between our model and existing state-of-the-art (SOTA) uni-modal driver action recognition methods using the Drive\&Act Dataset. 
Table~\ref{SOTA uni-modal table} presents the Top-1 and Mean-1 Accuracy (\%) for recognizing 34 fine-grained activities, comparing our work with SOTA approaches.
Our CM$^2$-Net model demonstrates superior performance across all uni-modalities. 
As illustrated in Table~\ref{SOTA uni-modal table}, CM$^2$-Net exhibits a significant improvement in both Top-1 and Mean-1 Accuracy. 
Specifically, in the RGB modality, CM$^2$-Net surpasses the top-performing TransDARC~\cite{transdarc} by 7.34\% in Top-1 and exceeds TSM~\cite{tsm} by 7.04\% in Mean-1. 
In the IR modality, our model shows 12.71\% and 5.16\% increases over TSM~\cite{tsm} in Top-1 and Mean-1 respectively.
Moreover, using Depth video input, CM$^2$-Net demonstrates a 13.39\% enhancement in Top-1 compared to the best SOTA models.
The notable performance in the IR and Depth validates the efficacy of ACMP in guiding their training.
Additionally, the results, particularly in the RGB input, underscore the effectiveness of our approach in leveraging semantic latent embedding for supervision.

\input{tabs/multi_modal}

To verify the effectiveness of cross-modal continual learning, we compare the multi-modal action recognition results with other fusion models in Table~\ref{SOTA multi-modal table}.
For a fair comparison, we just adopt the simple late fusion in our CM$^2$-Net.
In MDBU~\cite{mod4}, the authors present the optimal results achieved by using combinations of two modalities selected from eight input video streams in Drive\&Act.
It can be observed without employing complex fusion techniques, our CM2 model achieves a 7\% improvement in Top-1 and a 10\% improvement in Mean-1.
These results affirm that our original intention of enhancing the capabilities of each encoder in multi-modal action recognition is indeed a valid approach.

\vspace{-8pt}

\subsection{Ablation Study}

\vspace{-5pt}

\input{tabs/ablation3}

To demonstrate the efficacy of ACMP compared to existing methods, we take the same baseline UniFormerV2~\cite{uniformerv2} and test it on two previous methods.
In the first approach, we pre-train the baseline on each modality-specific dataset and then fine-tune it on the Drive\&Act dataset in the driving scenario.
To be more specific, the RGB encoder is pre-trained on K710~\cite{uniformerv2}, and the other two encoders are pre-trained on NTU RGB+D~\cite{nturgbd}.
For the second approach, we adopt the newly-tuned RGB encoder as the pre-trained backbone and then fine-tune it on the modality-specific data from Drive\&Act~\cite{driveact}.
We compare the results of these two methods with our ACMP on the IR and Dpeth modalities to verify the effectiveness of prompt from previous modalities.
As observed in Table~\ref{Ablation study 3}, the results of our prompt strategy surpass existing methods.
Compared to modality-specific pre-training, ACMP demonstrates significantly greater effectiveness in enhancing the performance of non-RGB encoders, even though NTU RGB+D is almost the biggest public action recognition dataset with IR and Depth input.
Meanwhile, the sub-optimal outcome observed by directly fine-tuning from the RGB encoder underscores a notable modality gap between RGB and other modalities. 
This suggests that the encoder may not be well-suited for extracting features from these alternative modalities.

\input{tabs/ablation4}

In the second part of the ablation study, we investigate the impact of prompting modalities in ACMP when training non-RGB modality encoders.
To be more specific, we conduct experiments on CM$^2$-Net with prompting from different modalities.
As indicated in Table~\ref{Ablation study 4}, the use of RGB prompt improves Mean-1 by more than 2\% in IR encoder and 7\% in the Depth encoder.
These results demonstrate that employing a pre-trained RGB encoder facilitates effective prompt transfer through cross-modal mapping in training other modal encoders. 
Furthermore, the inclusion of an IR prompt notably enhances the accuracy of the Depth encoder, affirming the feasibility of this cumulative approach.


\vspace{-8pt}

\subsection{Qualitative Analysis}

\vspace{-5pt}

\begin{figure}[t]
\centering
\includegraphics[width=0.48\textwidth]{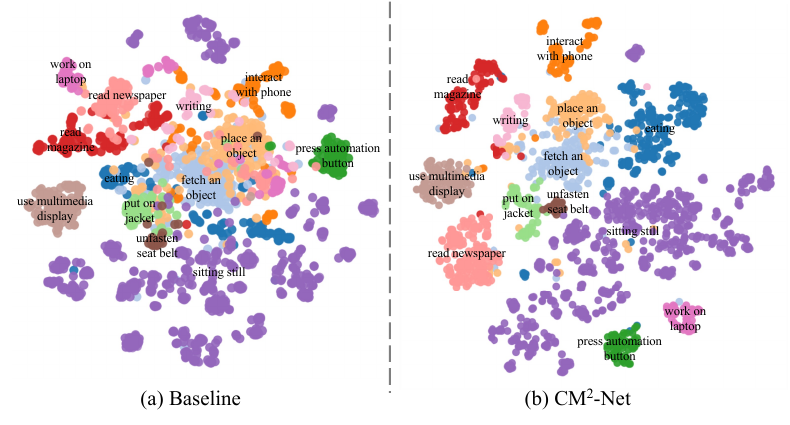}
\caption{
A t-SNE visualization of extracted features from different action categories. 
Different colors represent different actions. 
The features extracted by the baseline network~\cite{uniformerv2} (a) are more scattered than those extracted by CM$^2$-Net (b), which shows the efficacy of our method in mining discrimination information.
}
\label{visualize} \vspace{-10pt}
\end{figure}

To qualitatively demonstrate the representational learning capability of CM$^2$-Net, we utilize t-SNE~\cite{tsne} to visualize the Depth features extracted by both the baseline~\cite{uniformerv2} and our method, shown in Fig.~\ref{visualize}.
For the sake of simplicity and clarity in our visual analysis, we take the classes with a significant amount of samples.
Upon experiment, it is evident that the features extracted by the baseline network are scattered while the features extracted by CM$^2$-Net are compact and distinguishable.
This observation validates the superior capability of CM$^2$-Net in extracting discriminative features for non-RGB modalities.

%% file: tabs/uni_modal.tex
\begin{table}[t]
\caption{The benchmark results of CM$^2$-Net in comparison with existing single-modality methods on Drive\&Act.} \vspace{-15pt}
\begin{center}
\resizebox{0.4\textwidth}{!}{
\begin{tabular}{lccc}
\hline
Methods & Modality & Top-1 & Mean-1 \\ 
\hline
Pose~\cite{driveact} & 3D Skeleton & 55.17 & - \\
C3D~\cite{c3d} & NIR & 43.41 & - \\
P3D~\cite{p3d} & NIR & 45.32 & - \\
I3D~\cite{i3d} & NIR & 60.80 & - \\ 
\hline
I3D~\cite{i3d} & Depth & 60.52 & - \\
TSM~\cite{tsm} & Depth & \underline{63.76} & \underline{58.28} \\
\textbf{CM$^2$-Net(Ours)} & Depth & \textbf{77.15} & \textbf{63.91} \\
\hline
I3D~\cite{i3d} & IR & 64.98 & - \\
TSM~\cite{tsm} & IR & \underline{67.75} & \underline{59.81} \\
\textbf{CM$^2$-Net(Ours)} & IR & \textbf{80.46} & \textbf{64.97} \\
\hline
TSM~\cite{tsm} & RGB & 68.23 & \underline{62.72} \\
CTA-NET~\cite{ctanet} & RGB & 65.25 & - \\
TransDARC~\cite{transdarc} & RGB &  \underline{76.17} & 60.10 \\
\textbf{CM$^2$-Net(Ours)} & RGB & \textbf{83.51} & \textbf{69.76} \\
\hline
\end{tabular}
\label{SOTA uni-modal table}
}
\end{center} \vspace{-15pt}
\end{table}

%% file: tabs/multi_modal.tex
\begin{table}[t!]
\caption{The benchmark results of CM$^2$-Net in comparison with existing multi-modality methods on Drive\&Act.} \vspace{-15pt}
\begin{center}
\resizebox{0.4\textwidth}{!}{
\begin{tabular}{lccc}
\hline
Methods & Modality & Top-1 & Mean-1 \\ 
\hline
ResNet~\cite{resnet} & IR, Depth & 56.43 & 51.08 \\ 
TSM~\cite{tsm} & IR, Depth & 70.31 & 61.11 \\ 
MDBU~\cite{mod4} & Best 2 from all & \underline{76.91} & 62.02 \\ 
DFS~\cite{dar} & RGB, IR & 72.32 & \underline{62.87}\\
DFS~\cite{dar} & RGB, IR, Depth & 68.94 & 62.79\\
\textbf{CM$^2$-Net(Ours)} & RGB, IR, Depth & \textbf{83.92} & \textbf{72.10} \\
\hline
\end{tabular}
\label{SOTA multi-modal table}
}
\end{center} \vspace{-20pt}
\end{table}

%% file: tabs/ablation3.tex
\begin{table}[t!]
\caption{The comparison with the existing methods pre-trained on multi-modal dataset or adopt RGB encoder using the same baseline UniFormerV2. \textbf{PT}: Pre-training. \textbf{FT}: Fine-tuning.}\vspace{-10pt}
\begin{center}
\resizebox{0.48\textwidth}{!}{
\begin{tabular}{clcc}
\hline
Modality & Training Strategy & Top-1 & Mean-1 \\ 
\hline
\multirow{3}{*}{\textbf{IR}} 
& PT on NTU RGB+D~\cite{nturgbd}  & 72.56 & 59.64\\ 
& FT from RGB encoder & 76.39 & 61.40 \\
& \textbf{Prompt from RGB (ours)}  & \textbf{80.46} & \textbf{64.97} \\
\hline
\multirow{3}{*}{\textbf{Depth}}  
& PT on NTU RGB+D~\cite{nturgbd}  & 69.21 & 56.35\\ 
& FT from RGB encoder  & 72.93 & 59.35\\
& \textbf{Prompt from RGB \& IR (ours)} & \textbf{77.15} & \textbf{63.91} \\
\hline
\end{tabular}
}
\label{Ablation study 3}
\end{center} \vspace{-15pt}
\end{table}

%% file: tabs/ablation4.tex
\begin{table}[t!]
\caption{Ablation study the prompting modalities of ACMP.
\textbf{Note}: the prompts are only adopted in training.} \vspace{-10pt}
\begin{center}
\resizebox{0.48\textwidth}{!}{
\begin{tabular}{l|cc|cc}
\hline
\multirow{2}{*}{Modality} & \multicolumn{2}{c|}{Prompt} & \multirow{2}{*}{Top-1} & \multirow{2}{*}{Mean-1} \\
\cline{2-3} & RGB & IR \\
\hline
\textbf{IR} (w/o prompt)&&& 79.36 & 62.56 \\
\small  + RGB prompt & \checkmark && \textbf{80.46} & \textbf{64.97} \\
\hline
\textbf{Depth} (w/o prompt)&&& 73.89 & 54.85 \\
\small + RGB prompt & \checkmark && 76.30 & 62.28 \\
\small + IR prompt& \checkmark & \checkmark& \textbf{77.15} & \textbf{63.91} \\
\hline
\end{tabular}
\label{Ablation study 4}
}
\end{center} \vspace{-20pt}
\end{table}

%% file: Sec/5_conclusion.tex
\vspace{-10pt}

\section{Conclusion}

\vspace{-5pt}

In this work, we have developed a cross-modal driver action recognition model CM$^2$-Net, which is able to continually learn from new modalities to enhance the representation capability.
Furthermore, we proposed Accumulative Cross-modal Mapping Prompting (ACMP) to map the discriminative features learned from previous modalities into the newly-incoming modality feature space.
Then the new modality-specific encoders can be effectively trained with the prompt from the mapped features.
Consequently, this method facilitates the incorporation of knowledge from previously-learned modalities into the learning of new modalities.
Experimental results on the dataset Drive\&Act have shown that it can achieve state-of-the-art performance on both uni-modal and multi-modal driver action recognition.
\vspace{-10pt}
\section*{Acknowledgement}
\vspace{-5pt}
This study is supported under the RIE2020 Industry Alignment Fund - Industry Collaboration Projects (IAF-ICP) Funding Initiative, as well as cash and in-kind contribution from the industry partner(s).